\documentclass[11pt]{article}

\usepackage[truedimen,margin=25truemm]{geometry}

\usepackage{graphicx} 
\usepackage{epsfig} 
\usepackage{subfigure} 

\usepackage{amsmath}
\usepackage{amsthm}
\usepackage{amssymb}

\usepackage{algpseudocode}
\usepackage{algorithm}

\usepackage{color}

\usepackage{ulem}

\usepackage{booktabs}

\usepackage{hyperref}

\usepackage{graphicx} 
\usepackage{epsfig}
\usepackage{subfigure}

\usepackage{amssymb}
\usepackage{amsmath}
\usepackage{amsthm}
\usepackage{mathtools}
\usepackage[raggedright]{titlesec}

\def\vec#1{\mbox{\boldmath $#1$}}
\def\mat#1{\mbox{\bf #1}}

\usepackage{kasailab_def}

\title{Block-coordinate Frank-Wolfe algorithm and convergence analysis for semi-relaxed optimal transport problem}

\author{Takumi Fukunaga \thanks{Department of Communications and Computer Engineering, School of Fundamental Science and Engineering, WASEDA University, 3-4-1 Okubo, Shinjuku-ku, Tokyo 169-8555, Japan (e-mail: f\_takumi1997@suou.waseda.jp) } \and Hiroyuki Kasai \thanks{Department of Communications and Computer Engineering, School of Fundamental Science and Engineering, WASEDA University, 3-4-1 Okubo, Shinjuku-ku, Tokyo 169-8555, Japan (e-mail: hiroyuki.kasai@waseda.jp)}}

\begin{document}

\maketitle

\newcommand{\changeHK}[1]{\textcolor{red}{#1}}
\newcommand{\changeTF}[1]{\textcolor{blue}{#1}}
\newcommand{\note}[1]{\textcolor{magenta}{#1}}

\begin{abstract}

The optimal transport (OT) problem has been used widely for machine learning. It is necessary for computation of an OT problem to solve linear programming with tight mass-conservation constraints. These constraints prevent its application to large-scale problems. To address this issue, loosening such constraints enables us to propose the relaxed-OT method using a faster algorithm. This approach has demonstrated its effectiveness for applications. However, it remains slow. As a superior alternative, we propose a fast block-coordinate Frank--Wolfe (BCFW) algorithm for a convex semi-relaxed OT. Specifically, we {prove} their upper bounds of the worst convergence iterations, and equivalence between the linearization duality gap and the Lagrangian duality gap. Additionally, we develop two fast variants of the proposed BCFW. Numerical experiments have demonstrated that our proposed algorithms are effective for color transfer and surpass state-of-the-art algorithms. This report presents a short version of \cite{fukunaga2021fast}. The source code is available at \url{https://github.com/hiroyuki-kasai/srot}.

\end{abstract}
\begin{center}
{\small
{Published in IEEE ICASSP2022: https://doi.org/10.1109/ICASSP43922.2022.9746032 \cite{Fukunaga_ICASSP_2022}.}
}
\end{center}

\section{Introduction}
\label{Sec:Intro}
The optimal transport (OT) problem seeks an optimal {\it transport plan} or {\it transport matrix} by solving the total minimum transport cost from sources to destinations. It can express the distance between two probability distributions, which is known as Wasserstein distance \cite{Peyre_2019_OTBook}. This problem has been applied to widely diverse machine learning problems \cite{pmlr-v119-chen20e,Huang_arXiv_LCS_2020,Kasai_ICASSP_2020,fukunaga_ICPR_2020}. Among the OT problem formulations, the Kantorovich formulation is represented as convex linear programming (LP) \cite{Kantorovich_1942}. Thereby, many dedicated solvers such as an interior-point method can obtain the solutions. It is, nevertheless challenging to solve large-scale problems efficiently because its computational cost increases cubically in terms of the data size.

To address this issue, the Sinkhorn algorithm \cite{cuturi2013sinkhorn}, an {\it entropy-regularized} approach, has been proposed because it functions effectively and enables a parallel implementation. This faster computation derives from a differentiable and unconstrained convex optimization. Moreover, some stabler variants address overflow caused by small values of the regularizer, but they become slower \cite{chizat2018scaling}. Along another avenue of development, some works have reported that the strict constraints in the OT problem lead to performance degradation in some areas where mass is not necessarily preserved. This specific problem has been tackled recently by relaxing such tight constraints. A {\it constraint-relaxed} approach has attracted attention for use in various areas such as color transfer \cite{Rabin2014AdaptiveCT} and multi-label learning \cite{frogner2015learning}. Nevertheless, it remains slow.

To develop a faster solver generating sparser solutions for the OT problem and to work on its convex {\it semi-relaxed} formulation, this report is the first to describe a block-coordinate Frank--Wolfe (BCFW) algorithm with theoretical analysis. The FW algorithm is a variant of {\it projection-free} linear convex programming using a linear optimization oracle \cite{Frank_1956}. Thereby, an FW algorithm appropriate to large datasets has spread recently for the OT problem \cite{DiscreteSira2013,Rakotomamonjy_arXiv_2015,Courty_PAMI_2017,Paty_ICML_2019}. Furthermore, it can produce sparser solutions. However, because this algorithm requires solutions of all columns of the transport matrix at every iteration, its computational cost is not negligible for the very large matrix size. Therefore, we further incorporate a coordinate descent approach \cite{lacostejulien}, which updates one column randomly every iteration. The approach can achieve much smaller computational costs and faster convergence \cite{Wright_2015_MP}. Several papers have already described this method in the OT literature \cite{Perrot_NIPS_2016,Redko_NeurIPS_2020}, but its concrete theoretical properties for the relaxed OT problem have not been revealed. Consequently, the analyses presented in this paper can provide several fundamentally important theoretical contributions. The first contribution is to yield an upper-bound of the curvature constant without using a linear oracle, as in \cite{lacostejulien}. Then, we can obtain iteration complexities for $\epsilon$-optimality with FW and BCFW algorithms for the semi-relaxed OT problem. 
The second contribution is to show equivalence between {\it linearization duality gap}, a special case of the Fenchel duality gap and the Lagrangian duality gap. It becomes a certification of the current approximation values to monitor the convergence. We can use this property for the stopping condition in the proposed algorithm. The third contribution is to develop two fast variants of BCFW, i.e., the BCFW algorithms with pairwise-steps and away-steps. The final contribution is to demonstrate numerical evaluations of the color transfer problem, detailed analysis of the proposed BCFW, and its effectiveness including its faster variants in the semi-relaxed OT problem. Hereinafter, we designate vectors as bold lower-case letters $\vec{a},\vec{b},\vec{c},\cdots$ and present matrices as bold-face upper-case letters $\mat{A},\mat{B},\mat{C},\cdots$. 

\section{BCFW for semi-relaxed OT problem}
The evaluation in this paper specifically examines the semi-relaxed problem {\cite{blondel2018smooth}} with $\Phi(\vec{x},\vec{y})=\frac{1}{2\lambda}\|\vec{x}-\vec{y}\|_2^2$ because it is not only smooth but also convex. 
{This is formulated as}
\begin{equation}
\label{eq:EuculideanSquareRelaxedProblem}
\defmin_{\scriptsize \substack{\displaystyle{\mat{T}\geq \vec{0}},\\ \mat{T}^T\vec{1}_m = \vec{b}}}
\left\{f(\mat{T}):=\langle \mat{T},\mat{C} \rangle + \frac{1}{2\lambda}\|\mat{T}\vec{1}_n-\vec{a}\|_2^2\right\},
\end{equation}
where $\lambda$ is a {\it relaxation} parameter. The domain is transformed into {$\mathcal{M} = {b}_1\Delta_m \times {b}_2\Delta_m \times \cdots \times {b}_n \Delta_m$}, where ${b}_i\Delta_m$ represents the simplex of the summation ${b}_i$.

\subsection{{Proposed BCFW algorithm (Baseline)}}
\label{Sec:BaselineBCFW}
We first consider the FW algorithm for this problem; then we propose a faster block-coordinate Frank--Wolfe algorithm. 

\vspace*{0.1cm}

\noindent
{\bf Frank--Wolfe (FW) algorithm.} The gradient $\nabla f(\mat{T}) \in \mathbb{R}^{mn}$ is given as {$({\nabla_1 f(\mat{T})}^T,\dots,{\nabla_n f(\mat{T})}^T )^T$} where $\nabla_i f(\mat{T}) \in \mathbb{R}^{m}$ stands for the gradient on the $i$-th variable block ${b}_i\Delta_m$. The {linear subproblem} is equal to
\begin{equation}
	\label{eq:SemiRelaxedSubproblemImplement}
	\vec{s}_i = {b}_i\vec{e}_j = {b}_i\defargmin_{\scriptsize {\vec{e}_k \in \Delta_m ,k \in [m]}} \langle \vec{e}_k,\nabla_i f(\mat{T}^{(k)}) \rangle,
\end{equation}
where $\vec{e}_j$ represents the extreme point on probability simplex \cite{AwayStepClarkson}. The computational complexity of the subproblem can be improved greatly (\ref{eq:SemiRelaxedSubproblemImplement}). Its minute analysis is discussed in {\bf Section \ref{Sec:ComputationalComplexityAnalysis}}. After solving the solutions $\mat{S}=(\vec{s}_1, \vec{s}_2, \ldots, \vec{s}_n) \in \mathbb{R}^{m \times n}$, one must search an optimal step size $\gamma$. Classically, a {\it decay} stepsize (DEC) is often used in the FW algorithm, where $\gamma=2/(k+2)$ with the iteration number $k$. In this issue, a line-search algorithm can also be developed. 
Concretely, the quadraticity of the objective enables us to solve ${\min_{\gamma \in [0,1]}f((1-\gamma)\vec{x}+\gamma \vec{s})}$, and to calculate $\gamma$ directly.
\begin{algorithm}[t]
\caption{BCFW for semi-relaxed OT}   
\label{alg:BCFW-SROT}
\begin{algorithmic}[1]       
\Require{$\mat{T}^{(0)} = (\vec{t}^{(0)}_1,\dots,\vec{t}^{(0)}_n) \in {b}_1\Delta_m \times \cdots \times {b}_n\Delta_m$}
\For {$k=0 \dots K$}
\State Select index $i \in [n]$ randomly
\State Compute $\vec{s}_i = b_i \defargmin_{\scriptsize {\vec{e}_k \in \Delta_m}, k \in [m]}\ \langle \vec{e}_k,\nabla_if(\mat{T}^{(k)}) \rangle$

\State{Compute stepsize $\gamma$ }
\State {Update ${\vec{t}^{(k+1)}_i} = (1-\gamma)\vec{t}^{k}_i + \gamma \vec{s}_i$}
\EndFor
\end{algorithmic}
\end{algorithm}

\vspace*{0.1cm}

\noindent
{\bf Block-coordinate Frank--Wolfe (BCFW) algorithm.} We propose the block-coordinate Frank--Wolfe algorithm (BCFW) for the semi-relaxed problem converting the feasible set $\mathcal{M}$ into a Cartesian product. The algorithm procedure is shown in {\bf Algorithm \ref{alg:BCFW-SROT}}. It requires, at every iteration, solutions of the subproblem on the variable block selected randomly. Specifically, it is equivalent to (\ref{eq:SemiRelaxedSubproblemImplement}), but we solve the subproblem only for the $i$-th column, which is selected randomly. We adopt $\gamma=2n/(k+2n)$ as the stepsize, which is important for the convergence guarantee, as in {\bf Theorem \ref{thm:relaxedBCFWconvergence}}. Similarly to the FW algorithm, an exact line-search (ELS) algorithm can also be applied. The optimal stepsize $\gamma_{\rm LS}$ is calculated as 
\begin{equation}
\label{eq:SemiRelaxedBCFWStep}
\gamma_{\rm LS} = \frac{\displaystyle{\lambda\langle \vec{t}^{(k)}_i-\vec{s}_i,\vec{c}_i\rangle+\langle \vec{t}^{(k)}_i - \vec{s}_i,\mat{T}^{(k)}\vec{1}_n - \vec{a} \rangle}}{\displaystyle{ \|\vec{t}^{(k)}_i - \vec{s}_i\|^2}},
\end{equation}
where $\vec{t}_i$ represents the {$i$-th} column of \mat{T} and $\vec{s}_i$ denotes the solution of the $i$-th subproblem in (\ref{eq:SemiRelaxedSubproblemImplement}). As in {\bf Theorem \ref{thm:RelaxedFenchelDualityGap}}, the duality gap can be adopted as the stopping criterion. In a practical implementation, one can monitor the value of the duality gap because the subproblem is solved at every iteration. It is noteworthy that the BCFW algorithm can not compute the duality gap value from the solution of the value of (\ref{eq:SemiRelaxedSubproblemImplement}) because one must solve the subproblems on all the variable blocks. Therefore, computing the duality gap, if attempted every iteration, increases the runtime heavily. As a result, the benefit of the cheaper iteration complexity in BCFW is lost. Consequently, we monitor the duality gap every $n$ iteration, of which period is equal to that of the FW algorithm. 

Finally, we adopt two rules for sampling a column at each iteration: uniform random sampling and random permutation sampling. The former randomly selects $i \in [n]$, of which convergence analysis is given in {\bf Section \ref{Sec:TheoreticalResults}}. The latter runs a cyclic order on a permuted index. 
Those algorithms are represented, respectively, as BCFW-U and BCFW-P. Another sampling scheme using the duality gap is explained in {\cite{fukunaga2021fast}}.

\subsection{Theoretical results of the baseline BCFW algorithm}
\label{Sec:TheoreticalResults}
{\bf Convergence analysis.} We provide the worst convergence iteration of the proposed algorithm. 
\begin{Thm}
\label{thm:relaxedBCFWconvergence} 
Let $\mat{T}^*$ be the optimal solution of the semi-relaxed OT problem in (\ref{eq:EuculideanSquareRelaxedProblem}). Consider {\bf Algorithm \ref{alg:BCFW-SROT}} under the {arbitrary} initial point {$\mat{T}^{(0)}$} with a decay stepsize rule $k=\frac{2n}{k+2n}$. Then, one obtains $\mathbb{E}[f(\mat{T}^{({k})})] - f(\mat{T}^{*}) \leq \frac{2n}{k+2n}(C_f^\otimes+h_0)$, 
where $h_0=f(\mat{T}^{(0)})-f(\mat{T}^*)$, and where $C_f^\otimes$ is the curvature constant with $\leq \frac{4}{\lambda}$. Additionally, given an approximation precision constant $\epsilon$, if $\|\mat{C} \|_{\infty} \leq \frac{2}{\lambda}$, then {\bf Algorithm \ref{alg:BCFW-SROT}} requires at most the number of $\mathcal{O}(\frac{n}{\lambda \epsilon})$ for its convergence. Otherwise, it requires an additional number of $\frac{2nh_0}{\epsilon} \leq \frac{2n{\small \|\mat{C}\|_{\infty}}}{\epsilon}$.
\end{Thm}

For its proof, we first upper-bound the curvature constant $C_f^\otimes$, in consideration of the twice differentiability of $f(\mat{T})$ and the simplex structure. For arbitrary $\mat{T}^{(0)}$, we can also upper-bound $g(\mat{T}^{(0)})$ by $\| \mat{C}\|_{\infty}$. Finally, the upper-bound of the complexity is derived. It is particularly interesting that our proposed algorithm requires additional iterations when $\|\mat{C} \|_{\infty} > \frac{2}{\lambda}$. The full proof is given {in the \cite{fukunaga2021fast}}. 

\vspace*{0.2cm}
\noindent
{\bf Linearization duality gap and stopping criterion.} {\it Linearization duality} is a special case of the {\it Fenchel duality} in the FW algorithm \cite{lacostejulien,JaggiMartin2013}. Its duality gap can be calculated at the points $\vec{x}$ as $g(\vec{x})=\max_{\scriptsize \vec{s}' \in \mathcal{M}}\ \langle \vec{x} - \vec{s}', \nabla f(\vec{x}) \rangle = \langle \vec{x} - \vec{s}, \nabla f(\vec{x}) \rangle,$
where $\mathcal{M}$ is convex. It is noteworthy that adding $f(\vec{x})$ onto the linearization duality is equivalent to the {\it Wolfe duality} \cite{AwayStepClarkson}. For the semi-relaxed OT problem, we specifically give the equivalence between the linearization gap, denoted as $g(\mat{T})$, and the Lagrangian duality gap. 
\vspace*{-0.1cm}
\begin{Thm}
\label{thm:RelaxedFenchelDualityGap}
Consider the semi-relaxed problem in (\ref{eq:EuculideanSquareRelaxedProblem}). The linearization duality gap is provided as $g(\mat{T})= \langle \mat{T}-\mat{S},\mat{C}\rangle + \frac{1}{\lambda} \langle \mat{T}\vec{1}_n - \mat{S}\vec{1}_n,\mat{T}\vec{1}_n - \vec{a} \rangle,$ where \mat{S} is the solution of the subproblem (\ref{eq:SemiRelaxedSubproblemImplement}). Then, the linearization duality gap $g(\mat{T})$ is equivalent to the Lagrangian duality gap of {(\ref{eq:EuculideanSquareRelaxedProblem})}.
\end{Thm}
The full proof is presented {in \cite{fukunaga2021fast}}, but its proof sketch is the following:
We calculate the Lagrangian duality gap $g_L(\mat{T})$ directly as the difference the primal and dual function for the semi-relaxed problem. Finally, we prove the equivalence between $g_L(\mat{T})$ and $g(\mat{T})$ defined in this theorem. This theorem enables us to use $g(\mat{T})$ as both the linearization duality gap and the Lagrangian duality gap. Therefore, $g(\mat{T})$ is appropriate to the stopping criterion of the proposed algorithms. 

\subsection{{Fast variants of BCFW algorithm}}
\label{Sec:FastVariants}
In accordance with reports of earlier work \cite{WolfeBook1970,AwaystepMitchell,Osokin_ICML_2016}, this paper introduces the away-steps and the pairwise-steps into the proposed BCFW algorithm in an attempt to raise the convergence rate. 
First, we define the active set on the $i$-th ($i\in [n]$) variable block $\mathcal{S}_{i}$ as $\mathcal{S}_{i} = \lbrace \vec{e}_j \in \Delta_m : \alpha_{{\vec{e}_j}} > 0 \rbrace$, where $\alpha_{\vec{e}_j}$ represents the coefficient of the $j$-th extreme point $\vec{e}_j$. To remove the atoms, a new subproblem is defined as $\vec{v}_{i} = {\defargmin}_{\vec{v}' \in \mathcal{S}_i}\ \langle \vec{v}', \vec{c}_i + \frac{1}{\lambda}(\mat{T}\vec{1}_n - \vec{a} )\rangle.$ Similarly to the subproblem (\ref{eq:SemiRelaxedSubproblemImplement}), we can also solve this problem. Hereinafter, we define two directions as the {\it FW} direction $\vec{d}_{\rm FW}=\vec{s}_i - \vec{t}^{(k)}_i$ and the {\it Away} direction $\vec{d}_{\rm Away}=\vec{t}^{(k)}_i - \vec{v}_i $. Of those two directions, we adopt the one which decreases the value of the function. Subsequently, we calculate the stepsize $\gamma$ satisfying ${\min_{\gamma \in [0,\gamma_{\rm max}]}f((1-\gamma)\vec{x}+\gamma \vec{s})}$. This stepsize $\gamma_{\rm LS}$ is calculated by replacing $ \vec{t}^{(k)}_i-\vec{s}_i$ in (\ref{eq:SemiRelaxedBCFWStep}) with $\vec{d}$, where $\vec{d}$ is $\vec{d}_{\rm FW}$ or $\vec{d}_{\rm Away}$. Then, it is necessary to update both the selected column vector $\vec{t}^{(k)}_i$ and the selected active set $\mathcal{S}^{(k)}_i$. Similarly, we can construct the block-coordinate Pairwise Frank--Wolfe (BCPFW). In the BCPFW, only two atoms $\vec{s}_i$ and $\vec{v}_i$ are used. Setting the direction $\vec{d}_{\rm Pair} = \vec{s}_i - \vec{v}_i$ yields the movement between only two atoms and improvement of the convergence rate. Additional details are presented in \cite{fukunaga2021fast}.

\subsection{Computational complexity analysis}
\label{Sec:ComputationalComplexityAnalysis}
We present a complexity analysis of the semi-relaxed OT problem. The column vector of the transport matrix $\mat{T}$ is independent of other columns because the domain of the subproblem (\ref{eq:SemiRelaxedSubproblemImplement}) represents the Cartesian product of the probability simplex. Therefore, we can solve (\ref{eq:SemiRelaxedSubproblemImplement}) easily \cite{JaggiMartin2013} without solving LP directly. Thus, the total computational complexities {$\mathcal{O}((mn)^3\log (mn))$} are reduced to {$\mathcal{O}(mn)$}, which speeds up the time. Although the FW algorithm must call linear oracle for the $n$ variable block, the BCFW algorithm calls oracle only for one variable block selected randomly at every iteration. The convergence rate of the BCFW algorithm is equal to that of the FW algorithm. Results show that the computational complexities of the BCFW algorithm arise more quickly than those of the FW algorithm. Finally, as for the two fast variants, the complexity of the away-steps and that of pairwise-steps is $\mathcal{O}(|\mathcal{S}_i|)$ because their procedure is the same as that of the subproblem (\ref{eq:SemiRelaxedSubproblemImplement}). Because these algorithms also require the solution of the subproblem (\ref{eq:SemiRelaxedSubproblemImplement}), the complexity of those algorithms is equal to $\mathcal{O}(m+|\mathcal{S}_i|)$ at every iteration. Furthermore, because the inequality $1 \leq |\mathcal{S}_i| \leq m$ holds against its cardinality, the total computational complexities of BCAFW and BCPFW can be approximated by $\mathcal{O}(m)$.

\section{Numerical Evaluation}
\label{Sec:NumericalEvalations}
\vspace*{-0.1cm}
\noindent
{\bf Configuration.} This section first evaluates convergence behaviors of the baseline of the proposed BCFW in {\bf Section \ref{Sec:BaselineBCFW}}. Then, the comparisons among the fast variants of BCFW proposed in {\bf Section \ref{Sec:FastVariants}} are performed. Finally, we evaluate the color-transferred images visually using the baseline algorithm of BCFW. Note that, hereinafter, this section uses BCFW-U for the baseline algorithm unless stated otherwise. Algorithms are configured from the same initialization point $\mat{T}^{(0)}$, of which the first row is set $\vec{b}$. The algorithms are stopped when the iteration count reaches $1000$ epochs unless stated otherwise. We selected $\lambda= \{10^{-7},10^{-9}\}$ from our preliminary evaluations. All the experiments are executed on a 3.7 GHz Intel Core i5 CPU with 64 GB RAM. Finally, this experiment addresses the OT-based color transfer problem \cite{ColortransferOT} and uses two images, which are source image ``Gangshan District" by Boris Smokrovic, and the reference image ``Minnesota landscape arboretum" by Shannon Kunkle. 

\vspace*{0.1cm}

\noindent
\begin{figure}[t]
\begin{center}
	\begin{minipage}[t]{0.48\textwidth}
	\begin{center}
		\includegraphics[width=\textwidth]{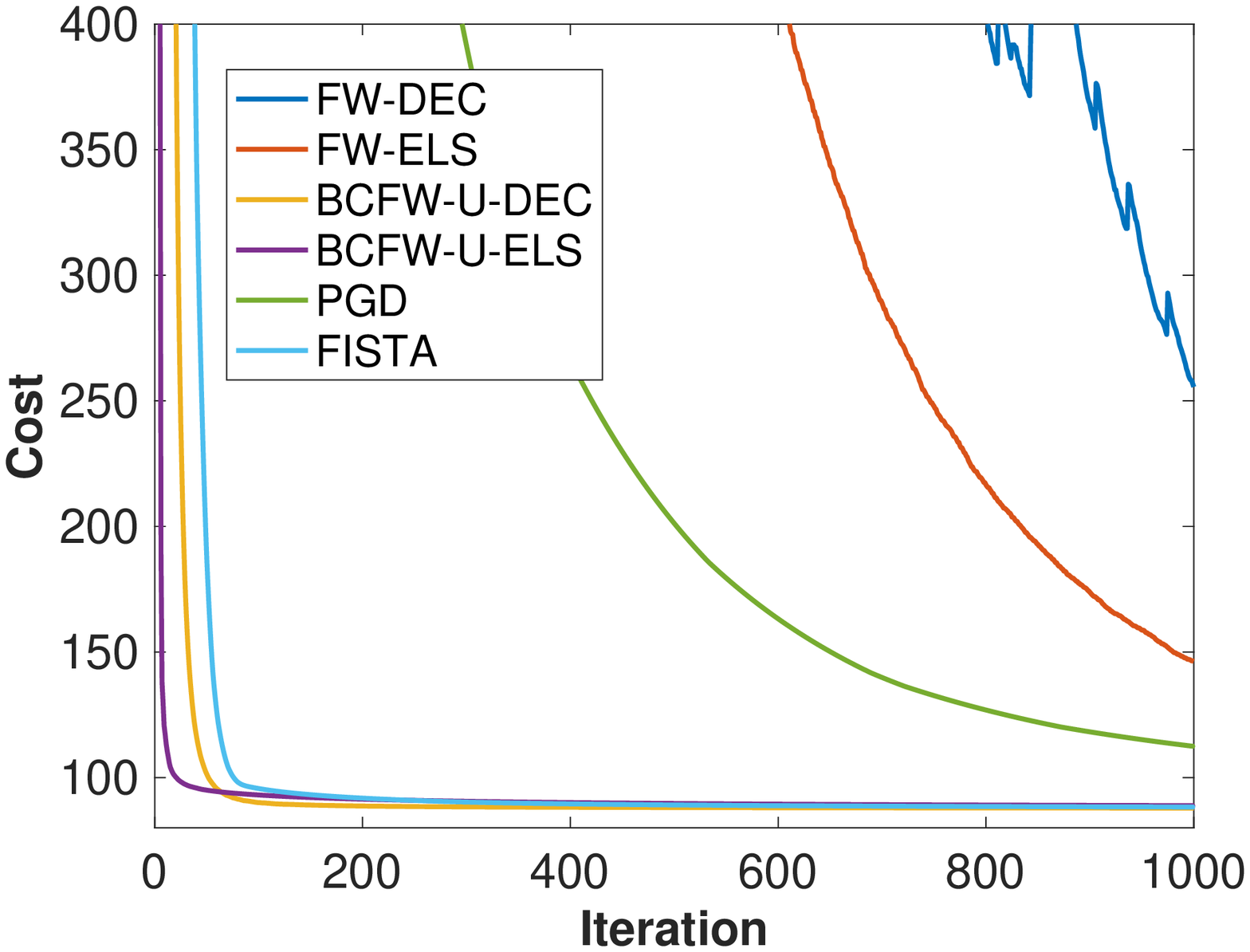}\\
		\vspace*{0.2cm}
		
		{(a) objective value : $f(\mat{T})$ }
		
	\end{center} 
	\end{minipage}
	\hspace*{0.2cm}
	\begin{minipage}[t]{0.47\textwidth}
	\begin{center}
		\includegraphics[width=\textwidth]{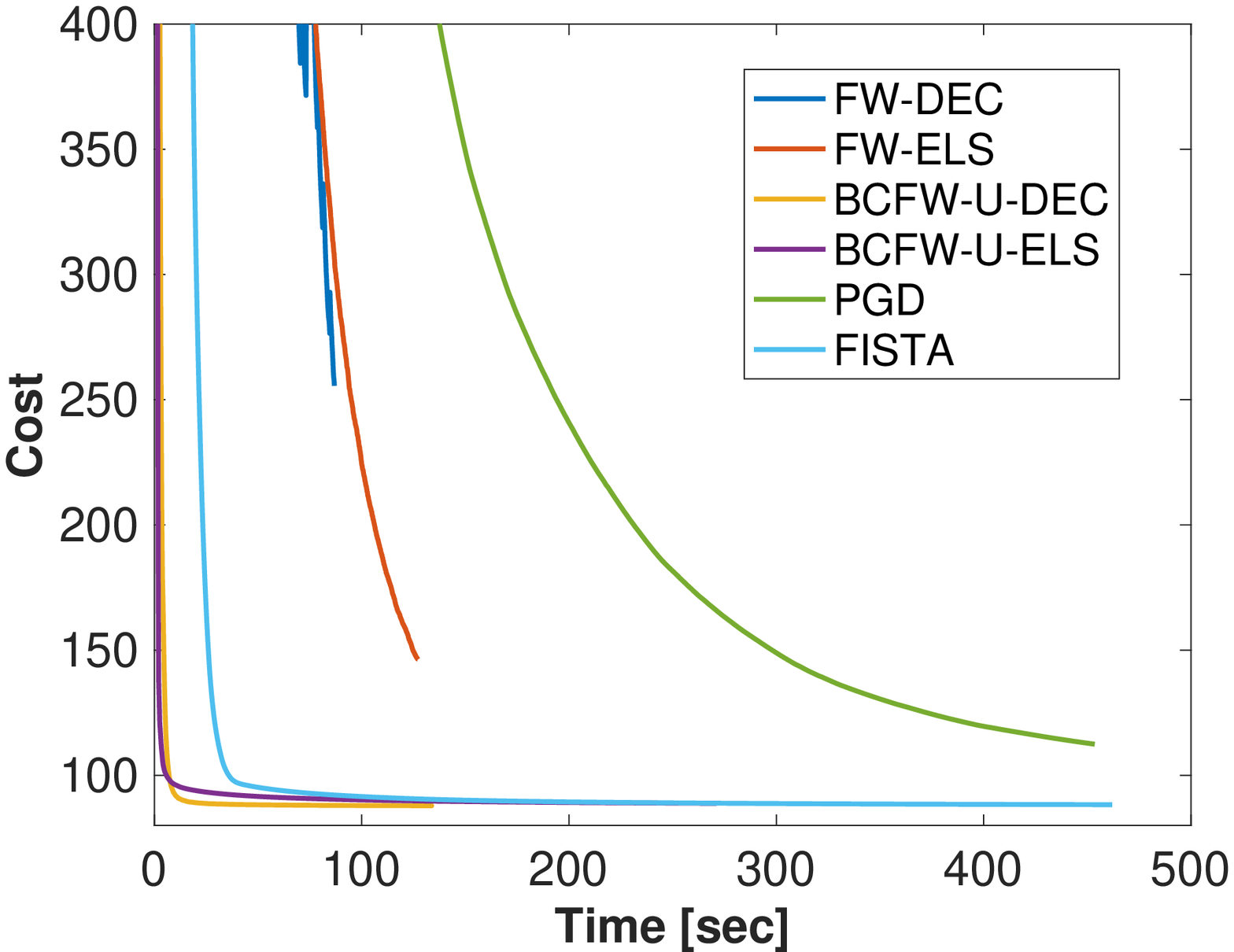}\\
		\vspace*{0.2cm}		
		
		{(b) objective value (time): $f(\mat{T})$ }
		
	\end{center} 
	\end{minipage}
	
	\vspace*{0.4cm}	
	
	\begin{minipage}[t]{0.48\textwidth}
	\begin{center}
		\includegraphics[width=\textwidth]{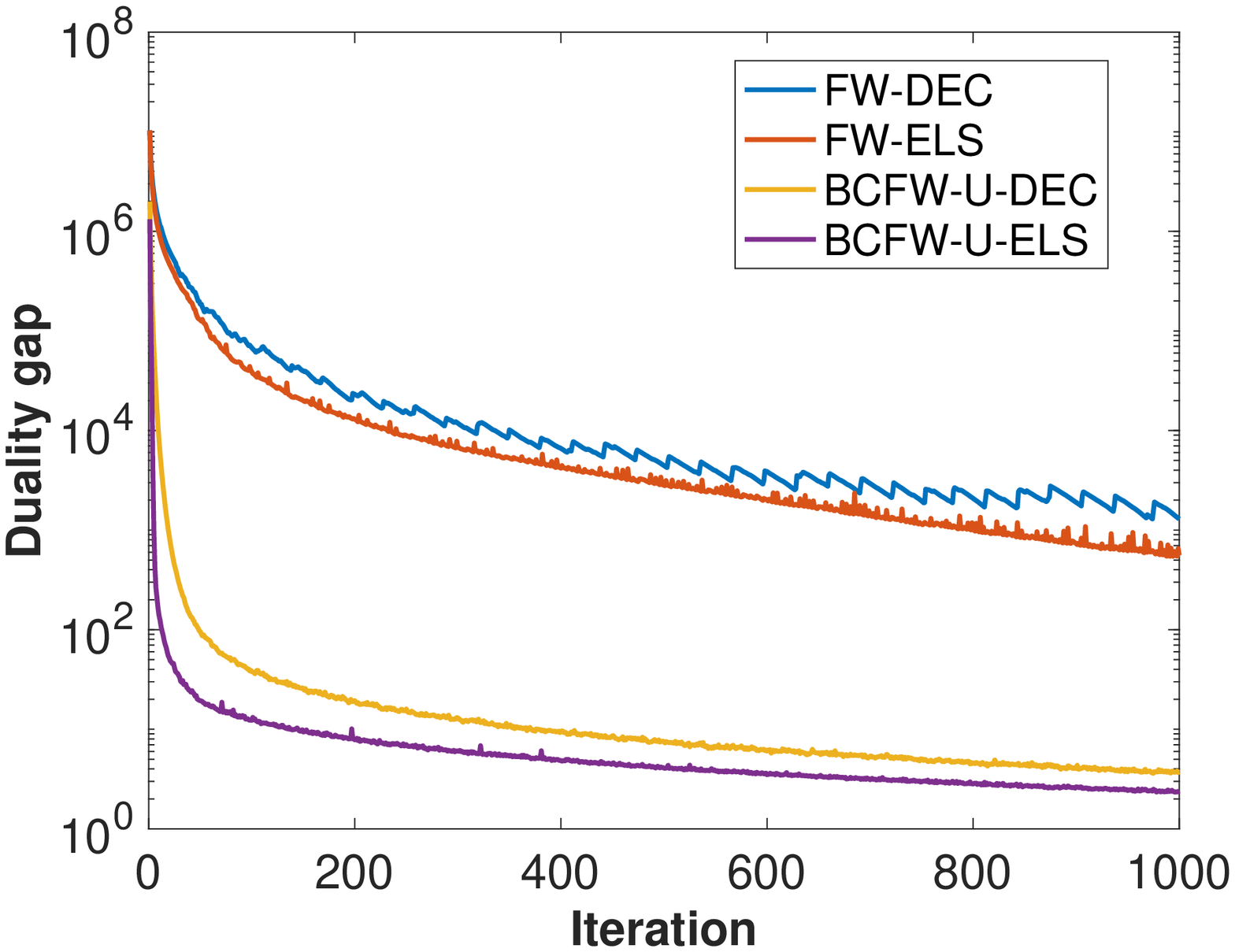}\\
		\vspace*{0.2cm}		
		
		{(c) duality gap : $g(\mat{T})$}
		
	\end{center} 
	\end{minipage}
	\hspace*{0.2cm}	
	\begin{minipage}[t]{0.475\textwidth}
	\begin{center}
		\includegraphics[width=\textwidth]{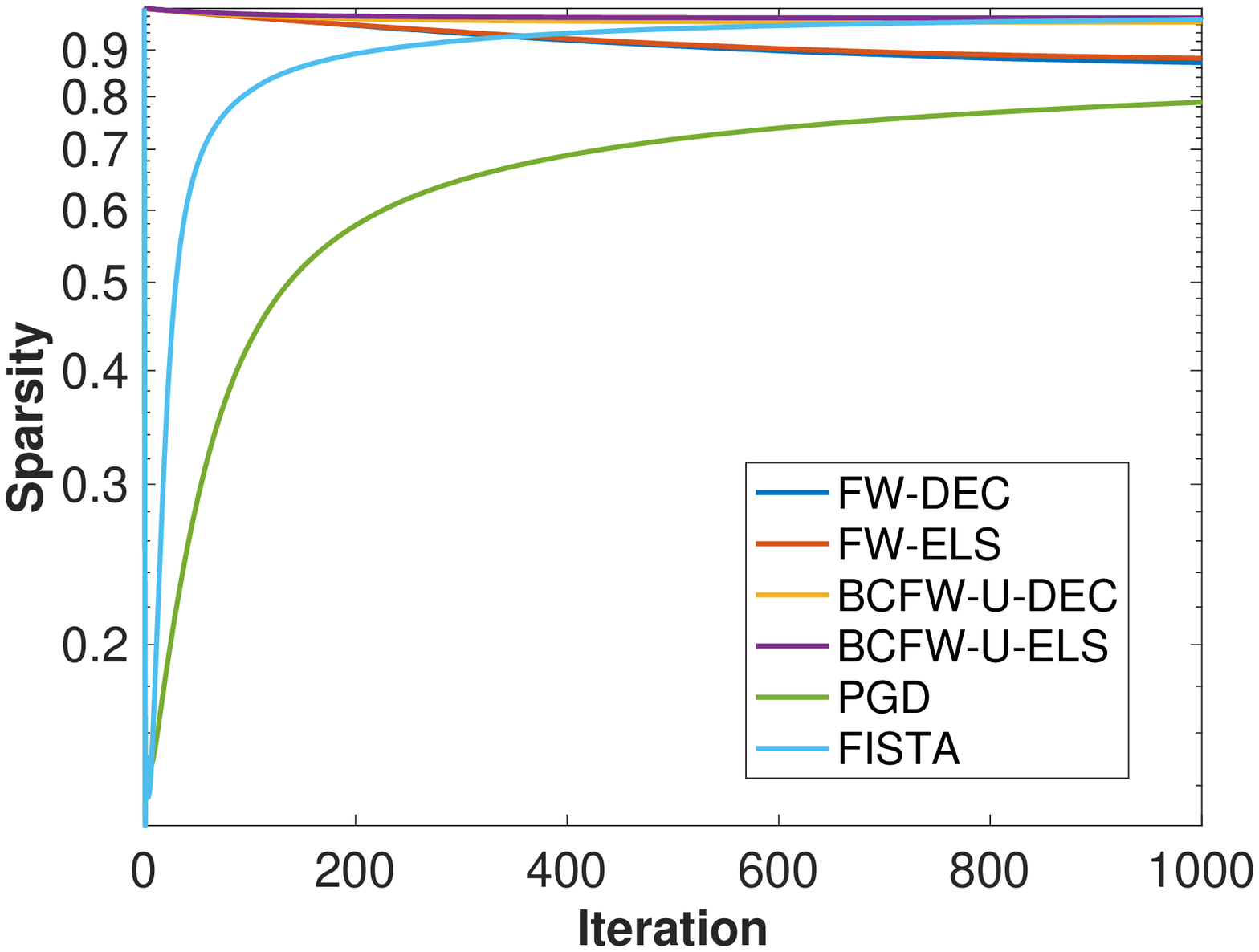}\\
		\vspace*{0.2cm}		
		
		{(d) sparsity}
		
	\end{center} 
	\end{minipage}
\caption{{Convergence of baseline BCFW and others}.}
\label{fig:ConvergencePerformance}
\vspace*{0.3cm}
\end{center}
\end{figure}
{\bf Evaluations of baseline BCFW.}
We compare our proposed algorithms with the projected gradient descent (PGD) method and the fast iterative shrinkage-thresholding algorithm (FISTA) \cite{Beck_2009_SIAMIS} for the semi-relaxed OT problem defined in (\ref{eq:EuculideanSquareRelaxedProblem}). We denote the corresponding FW and BCFW algorithms respectively as FW-DEC and FW-ELS, and BCFW-U-DEC and BCFW-U-ELS. We explain convergence performances in one specific case with $\lambda=10^{-7}$ and setting $m = n = 4096$. {\bf Fig.~\ref{fig:ConvergencePerformance}} presents the results. As the objective value and duality gap in {\bf Fig.~\ref{fig:ConvergencePerformance}}.(a)--(c) show, the BCFW-U algorithms are faster than the FW algorithm. In terms of iteration, the exact line-search stepsize rules outperform the decay stepsize rules in both the FW and BCFW-U algorithms. However, the decay stepsize is slightly more effective than the exact line-search in terms of computational time. Finally, because of the constructibility of the algorithms, we can be assured that our proposed algorithms output sparser solutions.

\vspace*{0.1cm}

\begin{figure}[tbp]
\begin{center}
	\begin{minipage}[t]{0.48\textwidth}
	\begin{center}
		\includegraphics[width=\textwidth]{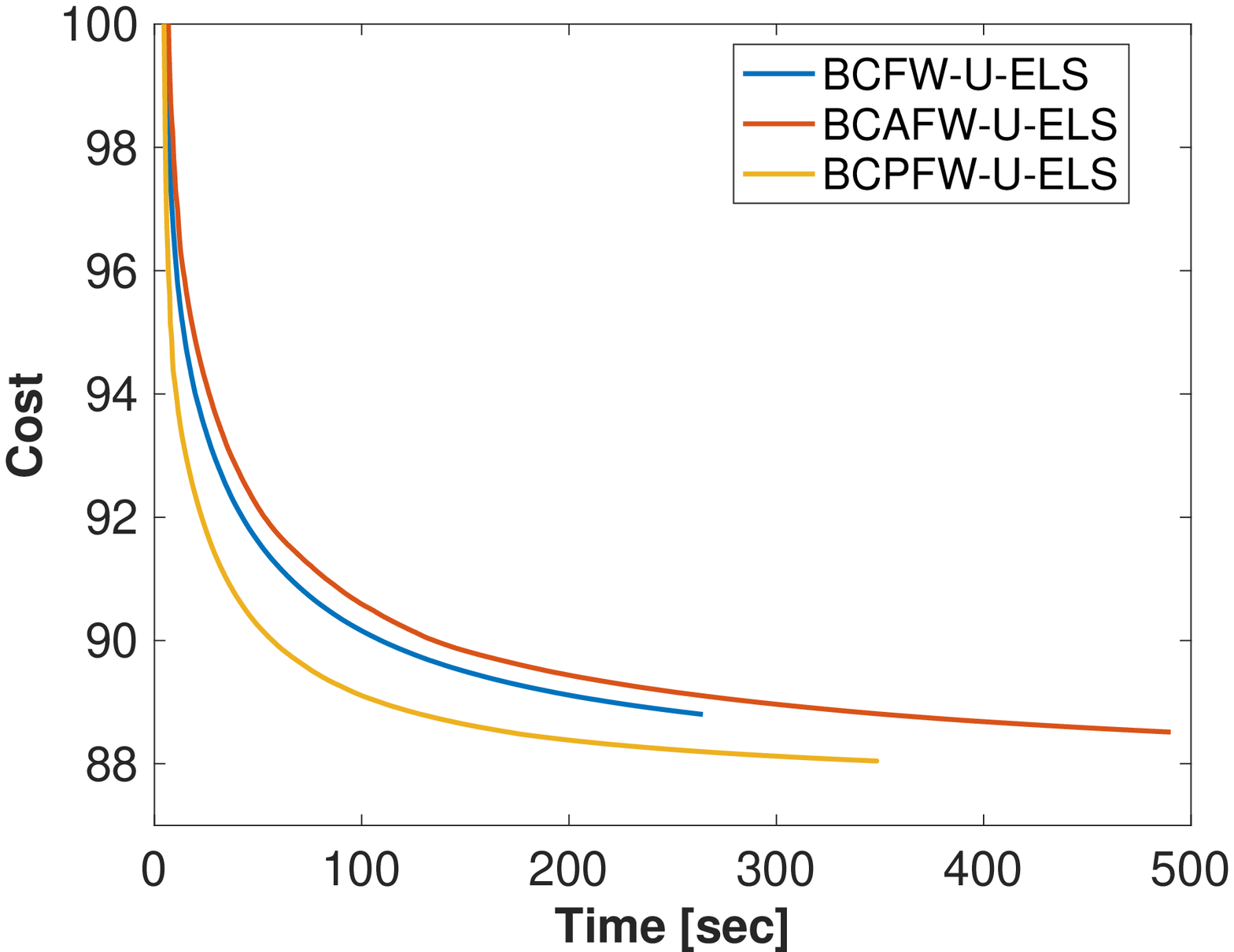}\\
		\vspace*{0.2cm}			
		
		{(a) objective value (time): $f(\mat{T})$ }
		
	\end{center} 
	\end{minipage}
	\hspace*{0.2cm}	
	\begin{minipage}[t]{0.485\textwidth}
	\begin{center}
		\includegraphics[width=\textwidth]{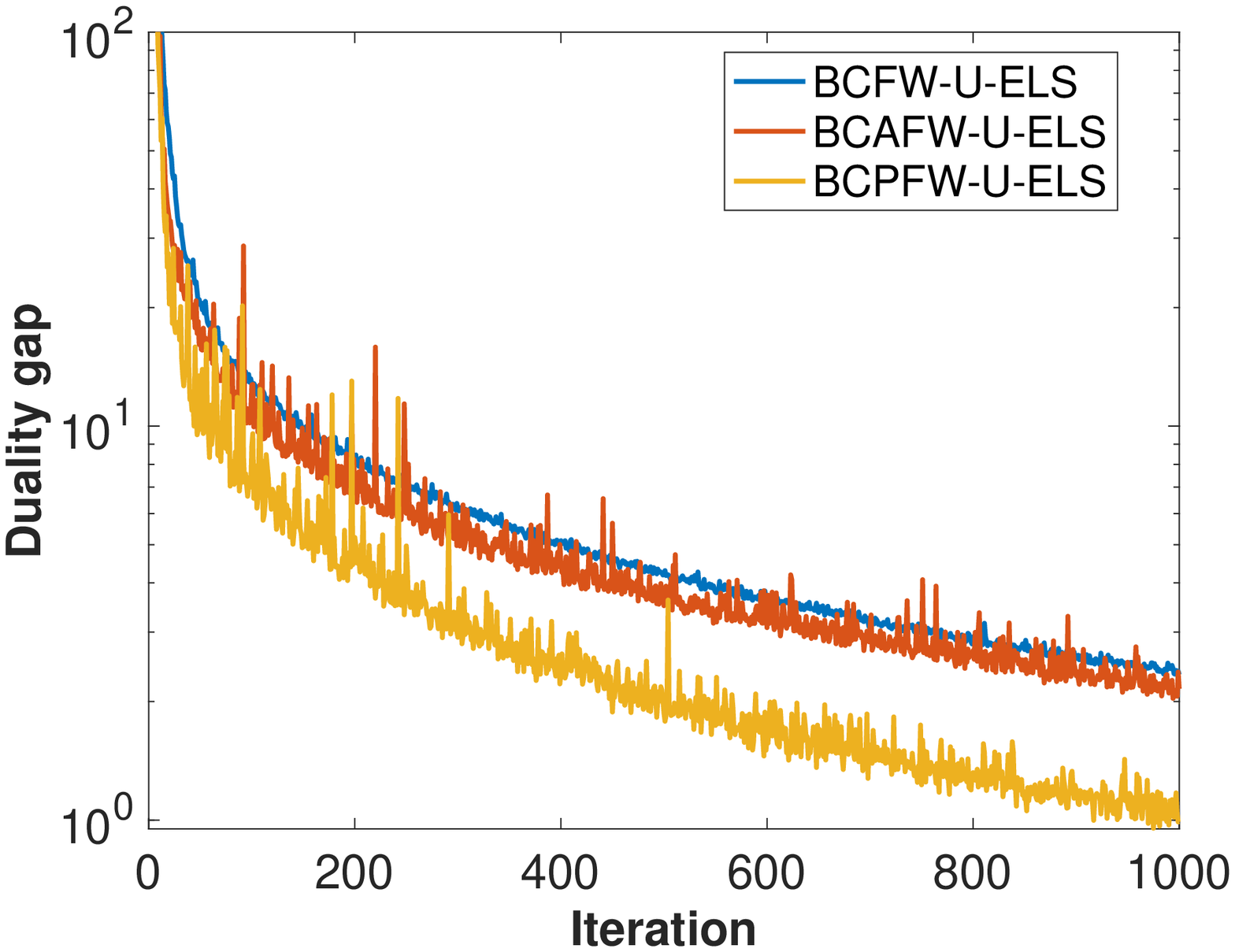}\\
		\vspace*{0.2cm}			
		
		{(b) duality gap : $g(\mat{T})$}
		
	\end{center} 
	\end{minipage}
\caption{{Convergence of fast variants: BCAFW and BCPFW}.}
\vspace*{-0.45cm}
\label{fig:FastVariantConvergencePerformance}
\end{center}
\end{figure}

\noindent
{\bf Evaluations of fast variants.}
This subsection presents evaluation of the performance improvements by two fast variants discussed in {\bf Section \ref{Sec:FastVariants}}, i.e., the BCFW with away-steps (BCAFW) and the BCFW with pairwise-steps (BCPFW). The experimental initializations are the same as the evaluation above. We explain the convergence performances when using the exact line-search stepsize rule (ELS), $\lambda=10^{-7}$ and $m = n = 4096$. {\bf Fig.~\ref{fig:FastVariantConvergencePerformance}} presents the results. From the duality gap in {\bf Fig.~\ref{fig:FastVariantConvergencePerformance}}.(a), one can understand that both the two fast variants BCAFW and BCPFW achieve faster convergence in terms of iteration. Also, BCPFW provides much better performances in terms of computational time.

\vspace*{0.1cm}

\begin{figure}[t]
\begin{center}
	\begin{minipage}[t]{0.49\textwidth}
	\begin{center}
		\includegraphics[width=\textwidth]{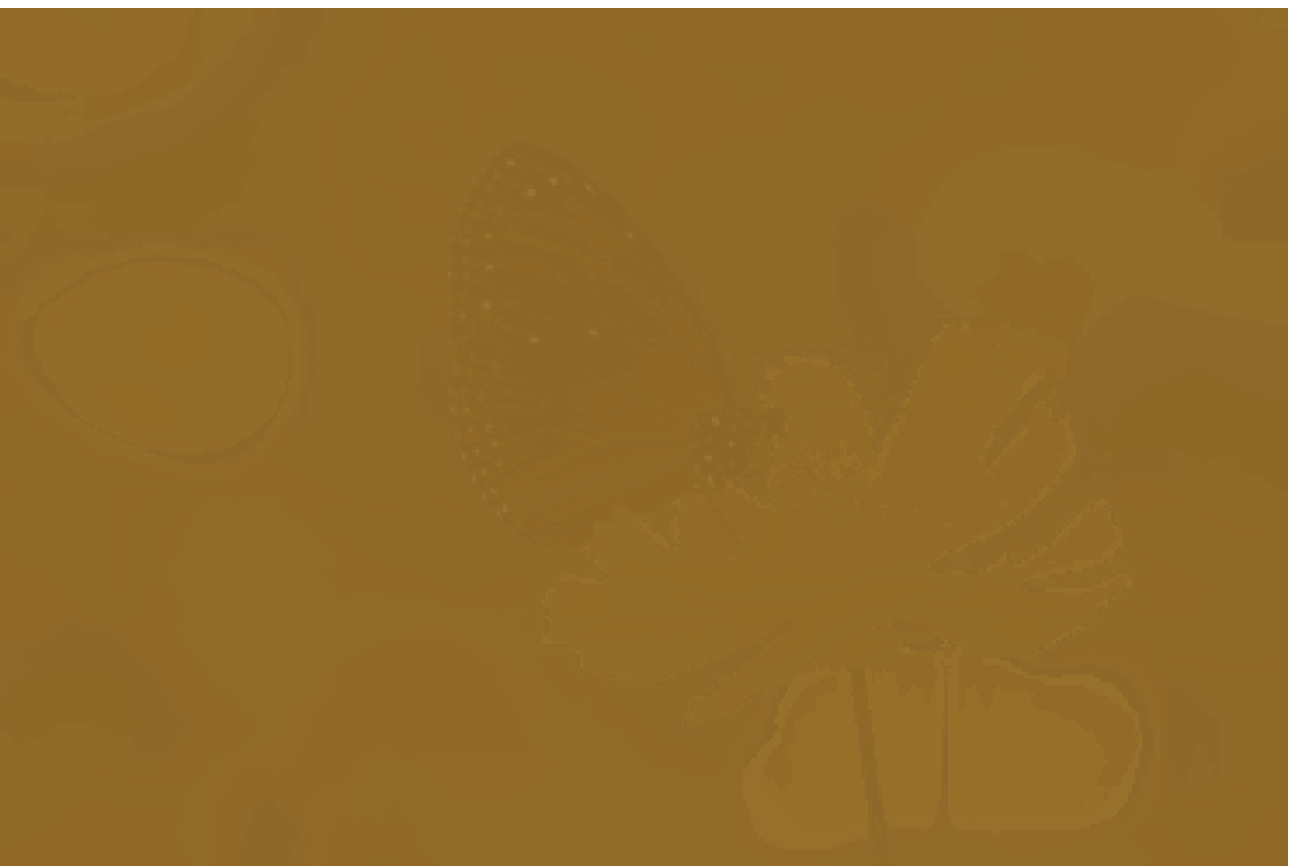}\\
				
		{$k = 3300$}		
	\end{center} 
	\end{minipage}
	\begin{minipage}[t]{0.49\textwidth}
	\begin{center}
		\includegraphics[width=\textwidth]{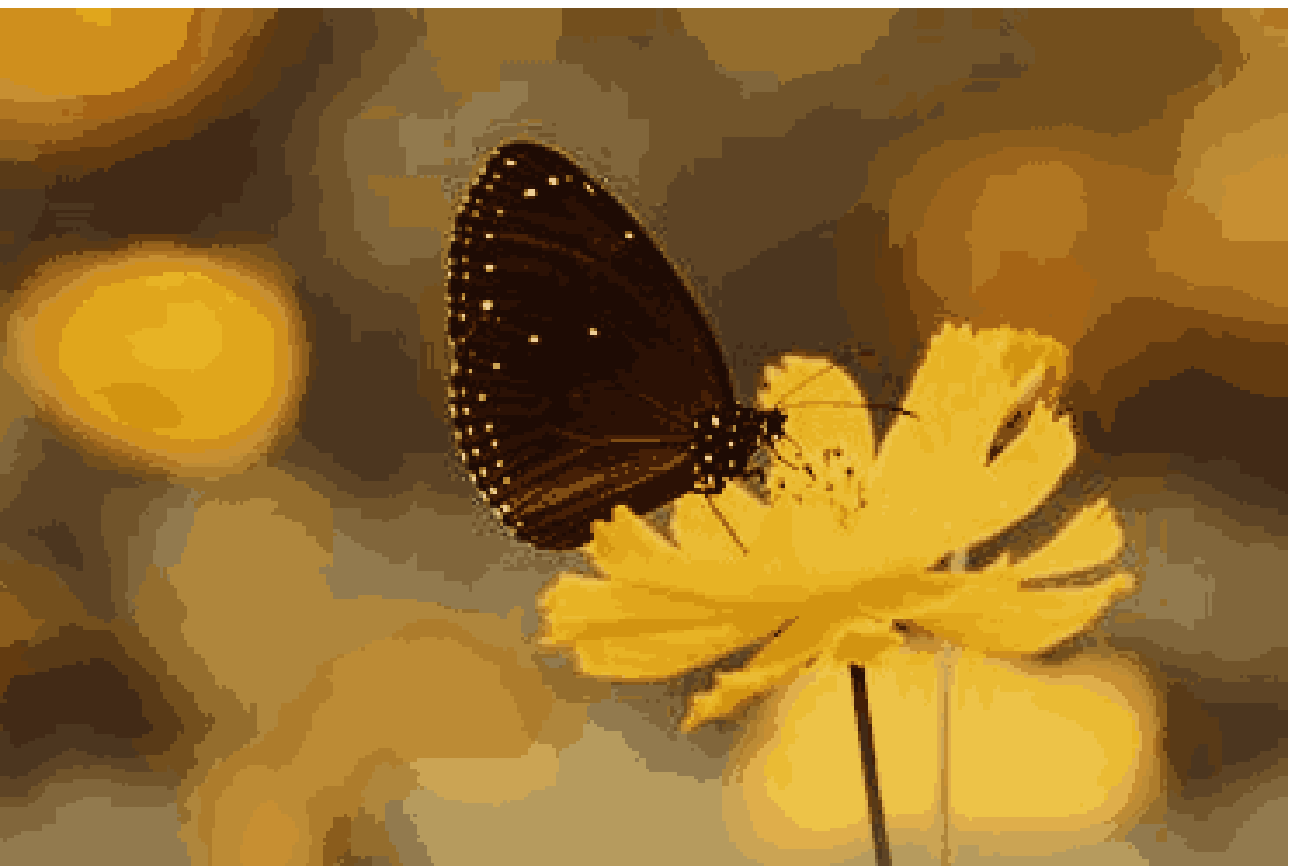}\\
					
		{$k = 10^6$}		
				
	\end{center} 
	\end{minipage}	
	\vspace*{0.4cm}
	
	{(a) transition of color-transferred images ($\lambda=10^{-9}$)}			
	\vspace*{0.4cm}
	
	\begin{minipage}[t]{0.49\textwidth}
	\begin{center}
		\includegraphics[width=\textwidth]{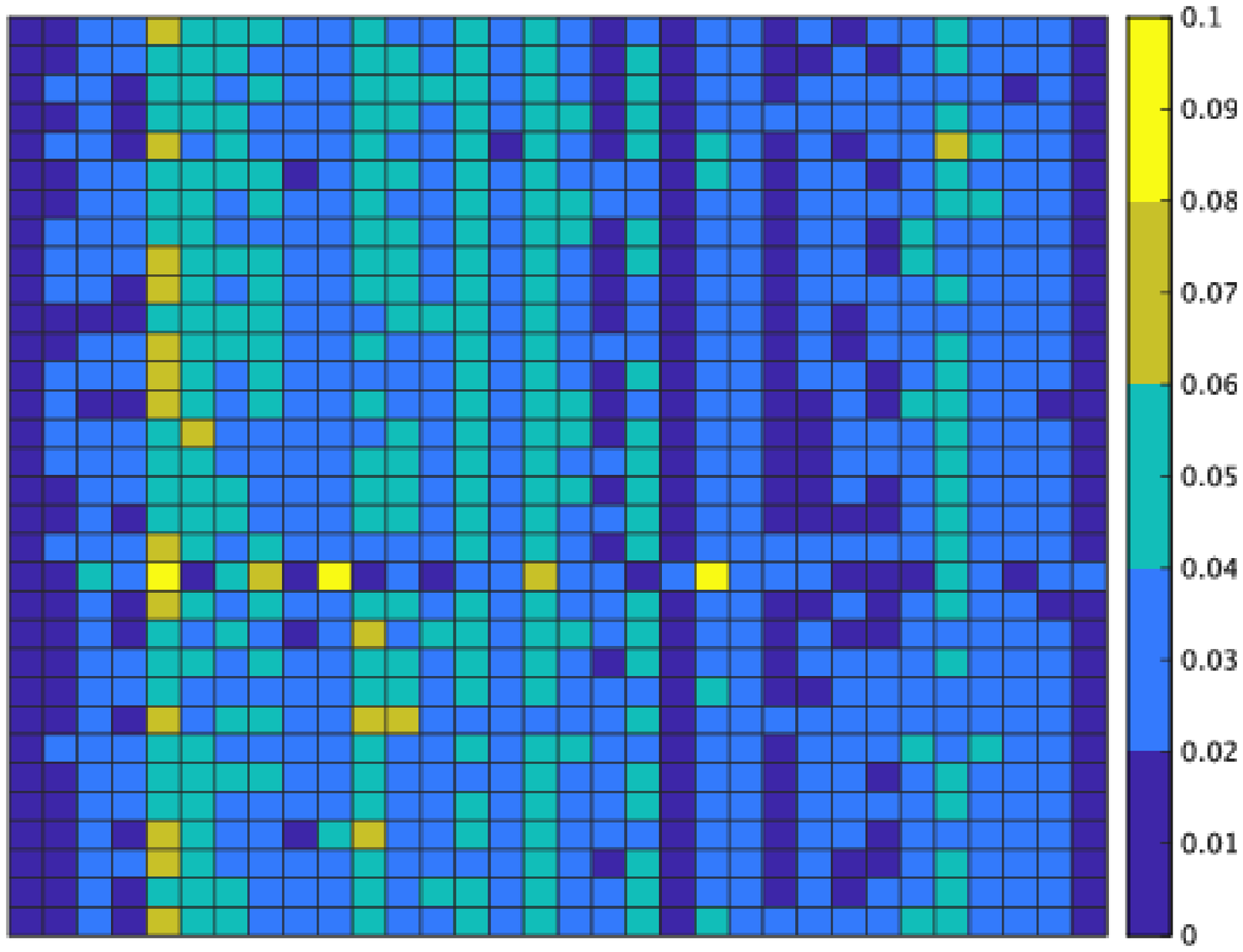}\\
				
		{$k = 3000$}
		
	\end{center} 
	\end{minipage}
	\begin{minipage}[t]{0.49\textwidth}
	\begin{center}
		\includegraphics[width=\textwidth]{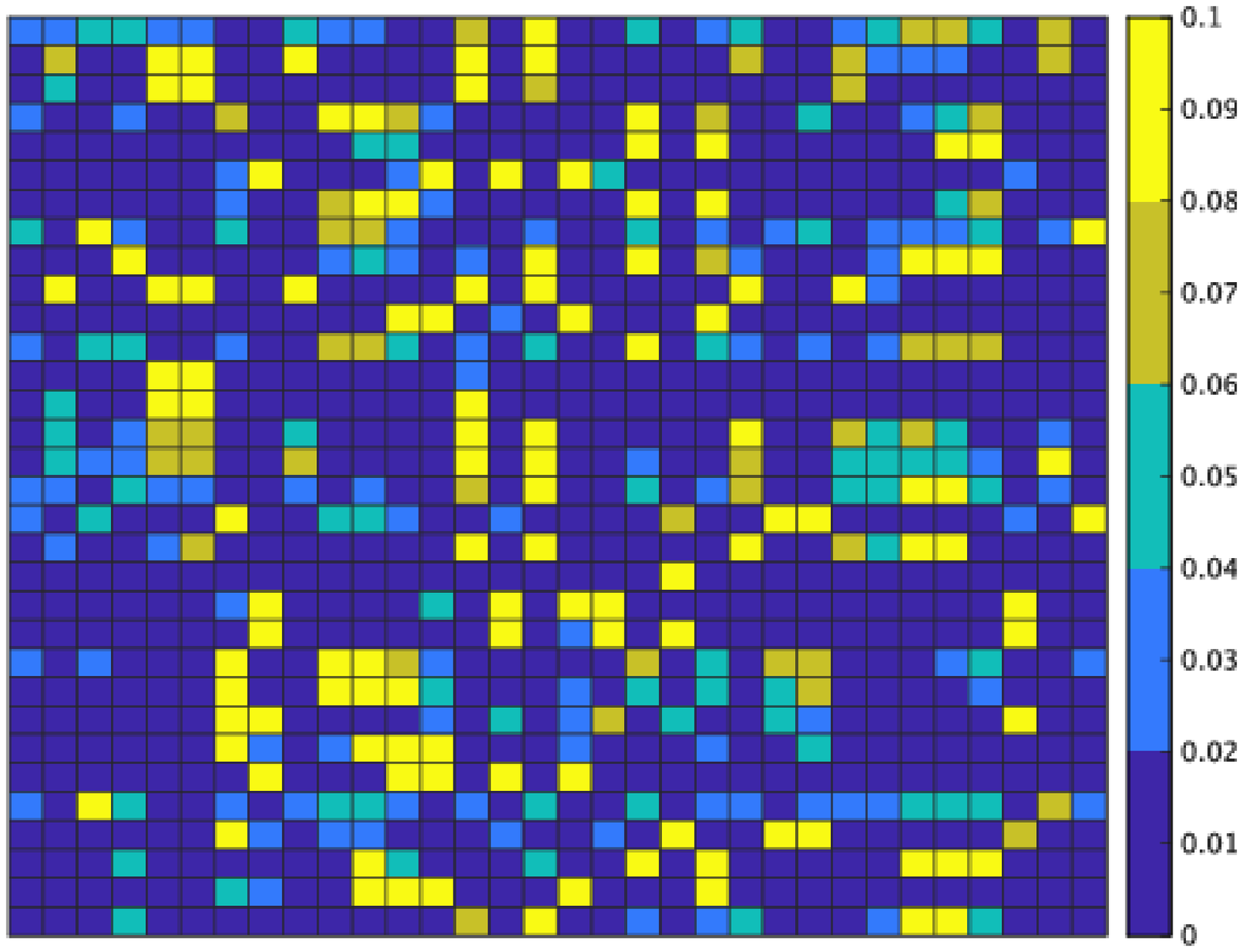}\\
				
		{$k = 10^6$}		
	\end{center} 
	\end{minipage}
	\vspace*{0.4cm}
	
	{(b) transition of the heat map of row-wise normalized transport matrices $\mat{T}^{(k)}$ ($\lambda=10^{-9}$)}

\centering
\caption{Transition of color-transferred images and heat-map of row-wise normalized transport matrices ($m=n=32$).}
\vspace*{-0.45cm}
\label{fig:CTImages_N32}
\end{center}
\end{figure}

\noindent
{\bf Evaluations of color-transformed images.}
This subsection presents specific evaluation of the effects of the relaxation parameter $\lambda$ on the color-transferred image using real-world images. A suitable setting is discussed to obtain images that are visually natural. We consider the case of $m\!=\!n\!=\!32$. The baseline BCFW with the decay stepsize rule is used: BCFW-U-DEC. {\bf Fig.~\ref{fig:CTImages_N32}} presents the results. The single color approaches the $\vec{b}$-weighed averaged color of the centroids of the reference image. Next, the single color is scattered. Naturally color-transferred images are generated. However, as the optimization process proceeds, the image at $k\!=\!10^6$ shows some artificial gray pixels in the background. {\bf Fig.~\ref{fig:CTImages_N32}}.(b) presents a heat map of the obtained transport matrices at the $k$-th iteration, $\mat{T}^{(k)}$. As a result of $\mat{T}^{(3000)}$, the matrix includes similar values along columns. In other words, the vertical blocks with similar colors can be recognized. However, this vertical structure disappears as the iteration proceeds.

\section{Conclusion}
After proposing BCFW for a convex semi-relaxed OT problem, we proved its upper-bounds of the worst convergence iterations, in addition to the equivalence between the linearization duality gap and the Lagrangian duality gap. Numerical evaluations demonstrated that our proposed algorithms outperform state-of-the-art algorithms across different settings.

\clearpage

\clearpage	
\normalem
\bibliographystyle{unsrt}
\bibliography{OptimalTransport,VariantFrankWolfe,others,Supplementary}

\begin{thebibliography}{10}

\bibitem{fukunaga2021fast}
T.~Fukunaga and H.~Kasai.
\newblock Fast block-coordinate {F}rank-{W}olfe algorithm for semi-relaxed
  optimal transport.
\newblock {\em arXiv preprint arXiv:2103.05857}, 2021.

\bibitem{Fukunaga_ICASSP_2022}
T.~Fukunaga and H.~Kasai.
\newblock Block-coordinate {F}rank--{W}olfe algorithm and convergence analysis
  for semi-relaxed optimal transport problem.
\newblock In {\em ICASSP}, 2022.

\bibitem{Peyre_2019_OTBook}
G.~Peyre and M.~Cuturi.
\newblock Computational optimal transport.
\newblock {\em Found. Trends Mach. Learn.}, 11(5-6):355--607, 2019.

\bibitem{pmlr-v119-chen20e}
L.~Chen, Z.~Gan, Y.~Cheng, L.~Li, L.~Carin, and J.~Liu.
\newblock Graph optimal transport for cross-domain alignment.
\newblock In {\em ICML}, 2020.

\bibitem{Huang_arXiv_LCS_2020}
J.~Huang, Z.~Fang, and H.~Kasai.
\newblock {LCS} graph kernel based on {W}asserstein distance in longest common
  subsequence metric space.
\newblock {\em Digit. Signal Process.}, 189:108281, 2021.

\bibitem{Kasai_ICASSP_2020}
H.~Kasai.
\newblock Multi-view {W}asserstein discriminant analysis with entropic
  regularized {W}asserstein distance.
\newblock In {\em ICASSP}, 2020.

\bibitem{fukunaga_ICPR_2020}
T.~Fukunaga and H.~Kasai.
\newblock {W}asserstein $k$-means with sparse simplex projection.
\newblock In {\em ICPR}, 2020.

\bibitem{Kantorovich_1942}
L.~Kantorovich.
\newblock On the transfer of masses (in russian).
\newblock {\em Dokl. Akad. Nauk}, 37(2):227--229, 1942.

\bibitem{cuturi2013sinkhorn}
M.~Cuturi.
\newblock {S}inkhorn distances: Lightspeed computation of optimal
  transportation distances.
\newblock In {\em NeurIPS}, 2013.

\bibitem{chizat2018scaling}
L.~Chizat, G.~Peyr{\'e}, B.~Schmitzer, and F.-X. Vialard.
\newblock Scaling algorithms for unbalanced optimal transport problems.
\newblock {\em Math. Comput.}, 87:2563--2609, 2018.

\bibitem{Rabin2014AdaptiveCT}
J.~Rabin, S.~Ferradans, and N.~Papadakis.
\newblock Adaptive color transfer with relaxed optimal transport.
\newblock {\em ICIP}, 2014.

\bibitem{frogner2015learning}
C.~Frogner, C.~Zhang, H.~Mobahi, M.~Araya-Polo, and T.~Poggio.
\newblock Learning with a {W}asserstein loss.
\newblock In {\em NeurIPS}, 2015.

\bibitem{Frank_1956}
M.~Frank and P.~Wolfe.
\newblock An algorithm for quadratic programming.
\newblock {\em Nav. Res. Logist. Q.}, 3:95--110, 1956.

\bibitem{DiscreteSira2013}
S.~Ferradans, N.~Papadakis, G.~Peyr{\'e}, and J.-F. Aujol.
\newblock Regularized discrete optimal transport.
\newblock {\em SIAM J. Imaging Sci.}, 7(3):1853--1882, 2013.

\bibitem{Rakotomamonjy_arXiv_2015}
A.~Rakotomamonjy, R.~Flamary, and N.~Courty.
\newblock Generalized conditional gradient: analysis of convergence and
  applications.
\newblock {\em arXiv preprint arXiv:1510.06567}, 2015.

\bibitem{Courty_PAMI_2017}
N.~Courty, R.~Flamary, D.~Tuia, and A.~Rakotomamonjy.
\newblock Optimal transport for domain adaptation.
\newblock {\em IEEE Trans. Pattern Anal. Mach. Intell.}, 39(9):1853--1865,
  2017.

\bibitem{Paty_ICML_2019}
F.-P. Paty and M.~Cuturi.
\newblock Subspace robust {W}asserstein distances.
\newblock In {\em ICML}, 2019.

\bibitem{lacostejulien}
S.~Lacoste-Julien, M.~Jaggi, M.~Schmidt, and P.~Pletscher.
\newblock Block-coordinate {F}rank-{W}olfe optimization for structural {SVMs}.
\newblock In {\em ICML}, 2013.

\bibitem{Wright_2015_MP}
S.~J. Wright.
\newblock Coordinate descent algorithms.
\newblock {\em Math. Program.}, 151:3--34, 2015.

\bibitem{Perrot_NIPS_2016}
M.~Perrot, N.~Courty, R.~Flamary, and A.~Habrard.
\newblock Mapping estimation for discrete optimal transport.
\newblock In {\em NeurIPS}, 2016.

\bibitem{Redko_NeurIPS_2020}
V.~Titouan, I.~Redko, R.~Flamary, and N.~Courty.
\newblock {CO}-optimal transport.
\newblock In {\em NeurIPS}, 2020.

\bibitem{blondel2018smooth}
M.~Blondel, V.~Seguy, and A.~Rolet.
\newblock Smooth and sparse optimal transport.
\newblock In {\em AISTATS}, 2018.

\bibitem{AwayStepClarkson}
K.~L. Clarkson.
\newblock Coresets, sparse greedy approximation, and the {F}rank-{W}olfe
  algorithm.
\newblock {\em ACM Trans. Algorithms}, 6(4):1--30, 2010.

\bibitem{JaggiMartin2013}
M.~Jaggi.
\newblock Revisiting {F}rank-{W}olfe: {P}rojection-free sparse convex
  optimization.
\newblock In {\em ICML}, 2013.

\bibitem{WolfeBook1970}
P.~Wolfe.
\newblock {\em Integer and nonlinear programming}.
\newblock Amsterdam : North-Holland Pub. Co., 1970.

\bibitem{AwaystepMitchell}
B.~F. Mitchell, V.~F. Dem'yanov, and V.~N. Malozemov.
\newblock Finding the point of a polyhedron closest to the origin.
\newblock {\em SIAM J. Control.}, 12(1):19--26, 1974.

\bibitem{Osokin_ICML_2016}
A.~Osokin, J.-B. Alayrac, I.~Lukasewitz, P.~Dokania, and S.~Lacoste-Julien.
\newblock Minding the gaps for block {F}rank-{W}olfe optimization of structured
  {SVM}s.
\newblock In {\em ICML}, 2016.

\bibitem{ColortransferOT}
F.~{Pitie} and A.~{Kokaram}.
\newblock The linear {M}onge-{K}antorovitch linear colour mapping for
  example-based colour transfer.
\newblock In {\em CVMP}, 2007.

\bibitem{Beck_2009_SIAMIS}
A.~Beck and M.~Teboulle.
\newblock A fast iterative shrinkage-thresholding algorithm for linear inverse
  problems.
\newblock {\em SIAM J. Imaging Sci.}, 2(1):182--202, 2009.

\end{thebibliography}

\end{document}